# Change Detection in Heterogeneous Optical and SAR Remote Sensing Images via Deep Homogeneous Feature Fusion

Xiao Jiang, Gang Li, *Senior Member, IEEE*, Yu Liu, Xiao-Ping Zhang, *Senior Member, IEEE*, and You He

*Abstract*—Change detection in heterogeneous remote sensing images is crucial for disaster damage assessment. Recent methods use homogeneous transformation, which transforms the heterogeneous optical and SAR remote sensing images into the same feature space, to achieve change detection. Such transformations mainly operate on the low-level feature space and may corrupt the semantic content, deteriorating the performance of change detection. To solve this problem, this paper presents a new homogeneous transformation model termed deep homogeneous feature fusion (DHFF) based on image style transfer (IST). Unlike the existing methods, the DHFF method segregates the semantic content and the style features in the heterogeneous images to perform homogeneous transformation. The separation of the semantic content and the style in homogeneous transformation prevents the corruption of image semantic content, especially in the regions of change. In this way, the detection performance is improved with accurate homogeneous transformation. Furthermore, we present a new iterative IST (IIST) strategy, where the cost function in each IST iteration measures and thus maximizes the feature homogeneity in additional new feature subspaces for change detection. After that, change detection is accomplished accurately on the original and the transformed images that are in the same feature space. Real remote sensing images acquired by SAR and optical satellites are utilized to evaluate the performance of the proposed method. The experiments demonstrate that the proposed DHFF method achieves significant improvement for change detection in heterogeneous optical and SAR remote sensing images, in terms of both accuracy rate and Kappa index.

## I. Introduction

CHANGE detection in remote sensing images is becoming increasingly important for rapid evaluation of natural disasters [1]. In many cases, the pre- and post-remote sensing images are collected by heterogeneous sensors. Among them, optical sensors and synthetic aperture radar (SAR) are the most commonly used. Optical sensors capture ground objects with high resolutions and multiple spectra [2][3], but their sensitivity to weather and sunlight conditions leads to difficulties of immediate acquisition of post-event qualified images [4]. In contrast, SAR is an active microwave sensor independent of weather and sunlight conditions, but it provides less information compared with optical sensors [5][6]. The complementary properties make them frequently used as a pair of pre-event monitoring (optical sensor) and rapid post-event acquisition (SAR) means [7]. Therefore, there exist strong needs for change detection in heterogeneous optical and SAR remote sensing images.

Change detection in heterogeneous remote sensing images is challenging due to their disparate feature representations of ground objects. It leads to infeasibility of direct comparisons (e.g., pixelwise difference and ratio) between heterogeneous images, which are commonly used for homogeneous images [8][9]. A number of methods have been proposed to address the issue. Jensen et al. introduce a post-classification comparison (PCC) method based on unsupervised clustering to detect wetland change in heterogeneous aircraft images [10]. In PCC, the pixels of the multi-temporal heterogeneous images are classified into different categories, such as wetland, forest, and rivers, to derive the corresponding classification maps. Then, the classification maps are compared to generate the regions of change. Mubea and Menz [11] later develop the PCC method by using support vector machine (SVM) for classification instead of unsupervised clustering. The performance of the PCC methods is susceptible to the classification accuracy and thus may be degenerated by the aggregation of classification errors [12]. Wu et al. [13][14] propose the Bayesian soft fusion framework by combining the classification results and the change detection probability to reduce the accumulation of misclassification errors on the homogeneous images. Different from the PCC-based methods, Niu et al. [15] and Volpi et al. [16] propose the joint-detection methods on the stacked multi-temporal heterogeneous images to avoid aggregated classification errors. Parts of the pixels of change and unchange in the stacked images are selected as the training samples. Although the joint-detection methods tend to achieve better performance than the PCC methods, extensive pixels/samples are required to learn the complicated relationship of the ground objects between heterogeneous images, which might be

Part of this work has been reported at IGARSS 2019 [30]. This work is supported in part by National Natural Science Foundation of China under Grants 61790551 and 61925106, and in part by the Civil Space Advance Research Program of China under Grant D010305. Corresponding author: Gang Li. Email: gangli@mail.tsinghua.edu.cn.

X. Jiang is with the Department of Electronic Engineering, Tsinghua University, Beijing 100084, China, and also with the Institute of Information Fusion, Naval Aeronautical University, Yantai 264001, China.

G. Li is with the Department of Electronic Engineering, Tsinghua University, Beijing 100084, China.

Y. Liu and Y. He are with the Institute of Information Fusion, Naval Aeronautical University, Yantai 264001, China.

X.-P. Zhang is with the Department of Electrical, Computer and Biomedical Engineering, Ryerson University, Toronto, ON M5B 2K3, CANADA.

inaccessible in practice [7].

Recent methods [7][17]-[26] based on homogenous transformation have achieved remarkable results with increasing popularity. Homogeneous transformation renders heterogeneous remote sensing images into the same feature space. Therefore, direct comparisons can be applied on the original and the transformed images with homogeneous features. Compared with the joint-detection methods, the methods based on homogeneous transformation do not need massive pixels/samples to learn the complicated relationship between the heterogeneous images [7][18][20]. Among these methods, Brunner et al. transform the pre-event optical images into SAR image space [17]. The estimated 3-D parameters of the landscapes from the optical satellite and the imaging parameters from SAR are utilized to generate the semantic content and the feature space of the transformed image, respectively. The change detection is then achieved on the transformed pre- and the original post-event SAR images. To avoid the employments of SAR imaging parameters, pixel transformation [18] and linear regression [19] are utilized to generate the feature space of the transformed images. The pixel transformation method [18] is later improved by transfer learning in [20]. Liu et al. propose a transfer classification method [21] for dealing with heterogeneous remote sensing data (e.g. SAR and optical images), and it can well manage the uncertain information by using multiple mapping value estimation strategy jointly with belief function theory during the transformation process. Gong et al. propose an unsupervised method by establishing the relationship between heterogeneous images via dictionary learning [22] and later develop a coupling convolutional neural network with iterative generation of detection results [7]. Kernel canonical correlation [23][24], manifold learning [25], and Bayesian nonparametric model [26] are also utilized to transform the heterogeneous images for change detection.

Among the above methods based on homogeneous transformation, there exists the problem that the features extracted for homogeneous transformation operate on the low-level space (e.g., pixel values [18]) and may corrupt the semantic content in the transformed results. The low-level features cannot describe accurately the image semantic content that is abstract in the high-level, especially in the regions with massive ground objects and complex scenes. This is because the low-level features offer limited capability for extraction of the image semantic content [28]. Therefore, the performance of homogeneous transformation is deteriorated, leading to inaccurate results of change detection.

Recent studies [28][29] on image style transfer (IST) based on deep convolutional neural networks (DCNN) [27] have received considerable attention. In IST, a natural image can be rendered into specific artistic styles from paintings. To achieve this, DCNN is used to separately extract the image semantic content and the style from the natural image and the painting, respectively. The final synthetic image is generated by using a cost function to combine the semantic content of the natural image and the style of the artistic painting.

The IST method aims to transfer the styles of natural images, but cannot meet the feature homogeneity for change detection. It uses a single cost function containing limited features to represent the image style, leading to feature inhomogeneity of the transformed image. The feature space is the feature set that represents the abstract semantic content in a specific image space. The style is a subset of the feature space with much less features. Both of them characterize the image semantic content, but the description of the style is much coarser than that of the feature space. For change detection, the feature spaces of the transformed and the heterogeneous images need to be the same, to make change detection feasible. As a result, the naïve IST method does not achieve the homogeneity of feature space for change detection in heterogeneous images.

In this paper, we present a new deep homogeneous feature fusion (DHFF) method for change detection in heterogeneous optical and SAR remote sensing images. In the proposed DHFF method, the homogeneous transformation that renders the heterogeneous images into the same feature space is considered as an IST problem. To the best of our knowledge, this is the first attempt to accommodate the concept of IST on change detection in heterogeneous remote sensing images.

The proposed DHFF method employs the DCNN that is used for IST to extract the semantic content and the style features, separately. Compared with the existing methods based on homogeneous transformation, the proposed method prevents the corruption of the semantic content by separate extraction, leading to accurate homogeneous transformation. Especially in the regions with multiple ground objects and complex scenes, the advancement is more evident because of the sufficient descriptions of the rich semantic content by the high-level features of DCNN.

To satisfy the feature homogeneity requirements of the transformation, we develop a new iterative IST (IIST) strategy. In the proposed IIST strategy, the cost function in each iteration measures the feature homogeneity in additional new feature subspaces, thus maximizing the feature homogeneity of the transformed image for change detection. Different from the naïve IST method using a cost function to measure the style in a single subspace with limited features, the cost function in the proposed method incorporates multiple cost functions to measure the feature homogeneity in additional new feature subspaces iteratively, leading to great improvement of feature homogeneity in the final transformed image. Randomized filter weights are employed to acquire additional new feature subspaces to enhance the description ability of the complete feature space. Based on the transformed image that achieves feature homogeneity by the new IIST strategy, the performance of homogeneous transformation and the accuracy of change detection are significantly improved.

In summary, the proposed method consists of the following key steps. First, the semantic content and the style features are separately derived from the heterogeneous optical and SAR remote sensing images by the high-level features of the DCNN originally designed for IST. Then, the IIST strategy is utilized to derive the transformed image with feature homogeneity. Finally, change detection is accomplished accurately on the original and the transformed images, both of which are in the homogeneous feature space.

Three datasets of optical and SAR remote sensing images are adopted to evaluate the performance of the proposed method. Among them, two datasets are acquired by GeoEye-1 (optical satellite) and RADARSAT-2 (SAR satellite). The third dataset consists of the optical and SAR images collected by Quickbird and COSMO-SkyMed satellites, respectively. The experiments demonstrate that the proposed DHFF model achieves significantly better accuracy rate and Kappa index than the existing change detection methods for heterogeneous optical and SAR remote sensing images, at the cost of the increased computational complexity.

The contributions of this paper are summarized as follows:

1) This work is the first attempt to apply the concept of IST for homogeneous transformation on the change detection task in heterogeneous remote sensing images. Different from the existing methods based on homogeneous transformation, the semantic content of the image is extracted separately by the DCNN with the high-level features, to avoid corruption and inaccurate change detection results.

2) Different from the naïve IST method that only transfers image styles, the proposed DHFF method measures and then achieves the feature homogeneity in additional new feature subspaces with the IIST strategy, to meet the requirements of feature homogeneity for change detection in homogeneous images.

The remainder of this paper is organized as follows. The change detection problem for heterogeneous optical and SAR remote sensing images is formulated based on DHFF in Section II. Section III describes the details of the proposed new method for change detection. The experimental results are presented in Section IV. Section V provides the concluding remarks.

## II. A New Model of Change Detection for Heterogeneous Optical and SAR Remote Sensing Images

### A. Problem Formulation

Assume that two heterogeneous remote sensing images, $I^{opt}$ and $I^{SAR}$, are available in a given region where an event of change happens. According to the properties of optical sensors and SAR mentioned above, $I^{opt}$ is assumed to be an optical image obtained before the change event happens (pre-event) while $I^{SAR}$ is a post-event intensity SAR image. Both of the two images are coregistered with each other. The objective of change detection is to find the regions of change from the heterogeneous optical and SAR images: $I^{opt}$ and $I^{SAR}$. In general, a binary map, named $BM$, revealing the final detected regions of change, is generated where the values "1" and "0" indicate the pixels of change and unchange, respectively.

### B. Deep Homogeneous Feature Fusion Framework

To detect the change between $I^{opt}$ and $I^{SAR}$, we propose a new homogeneous transformation framework incorporating the semantic content and the style features that is illustrated as follows:

$$BM = D(T_1(I^{opt}), T_2(I^{SAR})), \qquad (1)$$

where $T_1(\cdot)$ and $T_2(\cdot)$ are two homogeneous transformation functions; $D(\cdot)$ represents a change detection method that is commonly used for homogeneous images: $T_1(I^{opt})$ and $T_2(I^{SAR})$.

In the proposed framework (1), we choose the feature space of the optical image for homogeneous transformation.

Compared with SAR images, optical images are usually with higher resolutions and more details of ground objects. Transferring SAR images into optical image space will keep more semantic content in homogeneous transformation than transferring optical images into the feature space of SAR images. Therefore, we have:

$$T_1(I^{opt}) = I^{opt}. \qquad (2)$$

To transform $I^{SAR}$ into the optical image space, the concept of IST is applied to separately extract the semantic content and the style features of $I^{SAR}$ and $I^{opt}$, respectively. Then the transformed image is derived by the new IIST strategy to achieve the feature homogeneity:

$$T_2(I^{SAR}) = F(I^{opt}, I^{SAR}), \qquad (3)$$

where $F(\cdot)$ is the fusion operation to separately derive the semantic content and the style features of $I^{SAR}$ and $I^{opt}$, respectively, and then combine them by the IIST strategy.

## III. The Proposed Deep Homogeneous Feature Fusion (DHFF) Method

### A. Extraction Framework of Semantic Content and Style Features Based on DCNN

Before performing the separate feature extraction, the semantic content and the styles of the heterogeneous images should be defined. In the proposed method, the semantic content of an image is the semantic information of the ground objects (e.g., the types, shapes, and locations) that is maintained if captured by heterogeneous imaging sensors. The style of an image is the specific forms (e.g., textures) to describe and represent the ground objects, determined by different imaging sensors. The separate definitions of the semantic content and the style will help to avoid semantic content corruption in the following process of homogeneous transformation.

Fig. 1 shows the details of extraction of the features of the semantic content. The process of the style features is illustrated in Fig. 2. Similar to the naïve IST method in [28], the VGG network [33] is utilized as the framework to extract the semantic content and the style features.

As shown in Fig. 1, the layer hyperparameter Conv5-4 in the VGG network is selected and spanned into a vector as the semantic content features. In Appendix I, we explain the reason why Conv5-4 is selected.

The extraction of the style features is illustrated in Fig. 2. Similar to [28], the texture operator is applied on the spanned feature maps by the gram matrix. The style features are generated by the multi-scale layers of the VGG network, to provide thorough characterization of the image textures. In other words, the layers should cover all the scales of the network for complete descriptions.

Note that different from [28], the pooling layers covering all the scales are implemented instead of the ReLU layers because the pooling layers keep more useful texture information for style feature extraction [31][32]. The extracted textures from all the five pooling layers are concatenated to produce the style features $S(\cdot)$, as shown in Fig. 2(b).

Instead of the average pooling operation [28], the max pooling operation is employed in the VGG network to extract semantic content and styles, as shown in Figs. 1 and 2, respectively. In Appendix II, we demonstrate that the max pooling preserves the semantic content better compared with the average pooling.

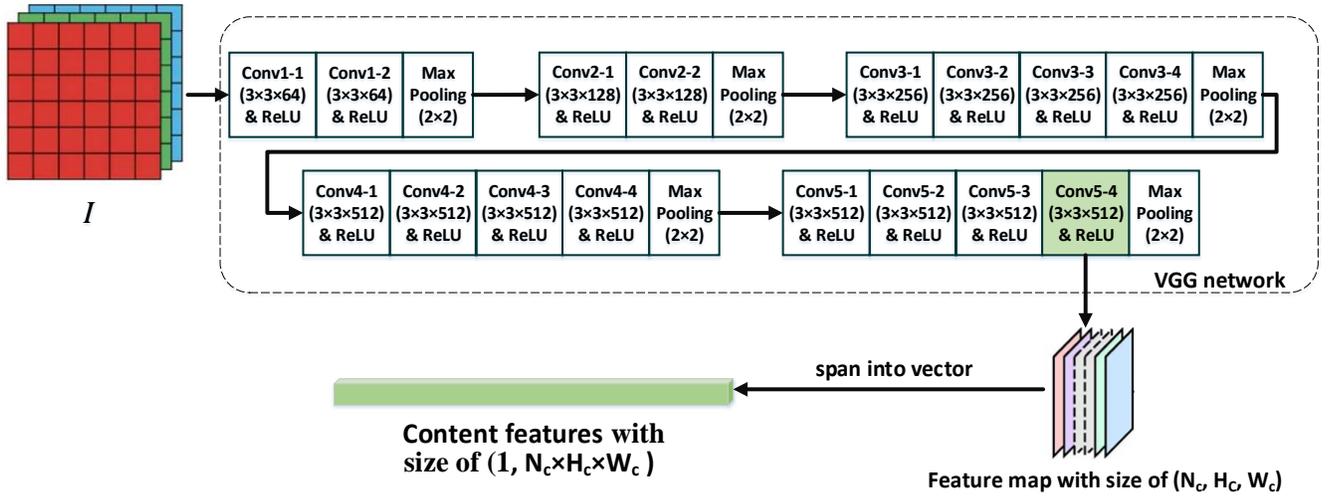

Fig. 1. Flowchart of the extraction of the semantic content features.

As can be seen in Figs. 1 and 2, entirely different features are extracted for the semantic content and the style of the image, separately. The output of the deepest convolutional layer Conv5-4 is employed as the semantic content features. The pooling layers covering all the scales of the image are combined with the texture operator to generate the style features. As a result, the semantic content is isolated from the style by the disparate features. Compared with the existing methods based on homogeneous transformation [7][17]-[26], the proposed method applies separate feature extraction by the DCNN (VGG network), capable of describing the high-level semantic content of the image with sufficiency and accuracy, especially in the regions with rich semantic content. Besides, the pooling layers with different scales can describe the image features represented by multi-scale textures. The semantic content and the style features carry distinct information of the image without confusion and represent sophisticate transformation relationship of multiple ground objects between the two heterogeneous images. Therefore, the semantic content of the original image is preserved without corruption, especially in the regions with multiple ground objects and complex scenes.

### B. New IIST Strategy Based on the VGG Network with Randomized Filter Weights

Here we aim to achieve the feature homogeneity $F(I^{opt}, I^{SAR})$ in (3), to derive the transformed image $T_2(I^{SAR})$, based on the extraction framework of the semantic content and the style features, as shown in Figs. 1 and 2.

We propose a new IIST strategy as follows:

$$T_2^k(I^{SAR}) = \arg\min_I L^k(I; I^{SAR}, I^{opt}) \quad k = 0, 1, 2, ..., \quad (4)$$

$$L^k(I; I^{SAR}, I^{opt}) = \lambda_c |C^k(I) - C^k(I^{SAR})|^2 + (1 - \lambda_c) |S^k(I) - S^k(I^{opt})|^2, \quad (5)$$

where $T_2^k(I^{SAR})$ is the updated transformed image generated by minimization of the cost function $L^k(\cdot)$ in the $k$-th IST iterations with $T_2^{k-1}(I^{SAR})$ employed as the initial image of the image solution $I$; $\lambda_c$ is the constant controlling the influence of the semantic content and the style features on the transformed image. For initialization, i.e., $k = 0$, $T_2^0(I^{SAR})$ is the output of the naïve IST method. It is generated in the feature subspace described by $C^0(\cdot)$ and $S^0(\cdot)$, of which the extraction framework is shown in Figs. 1 and 2 with the fixed pre-trained filter weights. The fixed filter weights of the extraction framework are pre-trained on the ImageNet [34]. For $k \geq 1$, $C^k(\cdot)$ and $S^k(\cdot)$ are added to measure the new feature subspace of homogeneity. In each iteration, the cost function $L^k(\cdot)$ is minimized by the limited-memory Broyden-Fletcher-Goldfarb-Shanno (L-BFGS) algorithm [28].

In each iteration, $T_2^k(I^{SAR})$ achieves the feature homogeneity of $C^k(\cdot)$ and $S^k(\cdot)$ by minimization of $L^k(\cdot)$ based on the initial image $T_2^{k-1}(I^{SAR})$, i.e., the minimization of $L^k(\cdot)$ serves as the transformation of $T_2^{k-1}(I^{SAR})$ along the additional new feature subspace represented by $C^k(\cdot)$ and $S^k(\cdot)$. Therefore, compared with $T_2^{k-1}(I^{SAR})$, $T_2^k(I^{SAR})$ achieves the feature homogeneity in the new feature subspace described by $C^k(\cdot)$ and $S^k(\cdot)$, in addition to the feature homogeneity achieved in the feature subspace described by $C^{k-1}(\cdot)$ and $S^{k-1}(\cdot)$. In this way, $T_2^k(I^{SAR})$ achieves the feature homogeneity in the feature subspaces described by the semantic content features $C^0(\cdot)$, $C^1(\cdot)$,… $C^k(\cdot)$ and the style features $S^0(\cdot), S^1(\cdot),… S^k(\cdot)$. In other words, $T_2^k(I^{SAR})$ is refined by the new additional feature subspace described by $C^k(\cdot)$ and $S^k(\cdot)$.

To extract the features $C^k(\cdot)$ and $S^k(\cdot)$, $k \geq 1$, effectively, the filter weights of the convolutional layers in the extraction framework shown in Figs. 1 and 2 are randomized in each loop of the iterations. Assume the filter weights of the pre-trained VGG network that derive $C^0(\cdot)$ and $S^0(\cdot)$ as $\mathbf{W}_i^0 = \{w_{i1}^0, w_{i2}^0, ..., w_{in}^0\}$, $i = 1, 2, ..., 16$ that includes all the $n$ weight values of the $i$-th convolutional layer in the pre-trained VGG network with 16 convolutional layers. The filter weights to derive $C^k(\cdot)$ and $S^k(\cdot)$, $k \geq 1$, are given by:

$$\mathbf{W}_i^k = \mathbf{W}_i^0 + \alpha_i \cdot \mathbf{X}_i^k, \quad (6)$$

where $\mathbf{W}_i^k$ indicates the filter weights of the $i$-th convolutional layer in the $k$-th iteration, $\alpha_i$ is a constant controlling the intensity of the randomization of the $i$-th convolutional layer, $\mathbf{X}_i^k = \{x_{i1}^k, x_{i2}^k, \dots, x_{in}^k\}$ represents $n$ i.i.d. Gaussian variables derived in the $k$-th iteration. For each variable, $x_{ij}^k \sim N\left(0, \text{Var}(\mathbf{W}_i^0)\right)$ with $\text{Var}(\mathbf{W}_i^0) = \frac{1}{n-1}\sum_{j=1}^n (w_{ij}^0 - \overline{w_i^0})^2$, as the estimated variance of $\mathbf{W}_i^0$ and $\overline{w_i^0} = \frac{1}{n}\sum_{j=1}^n w_{ij}^0$. The Gaussian randomization is to assure the common assumption of normal distribution of the convolutional layer weights.

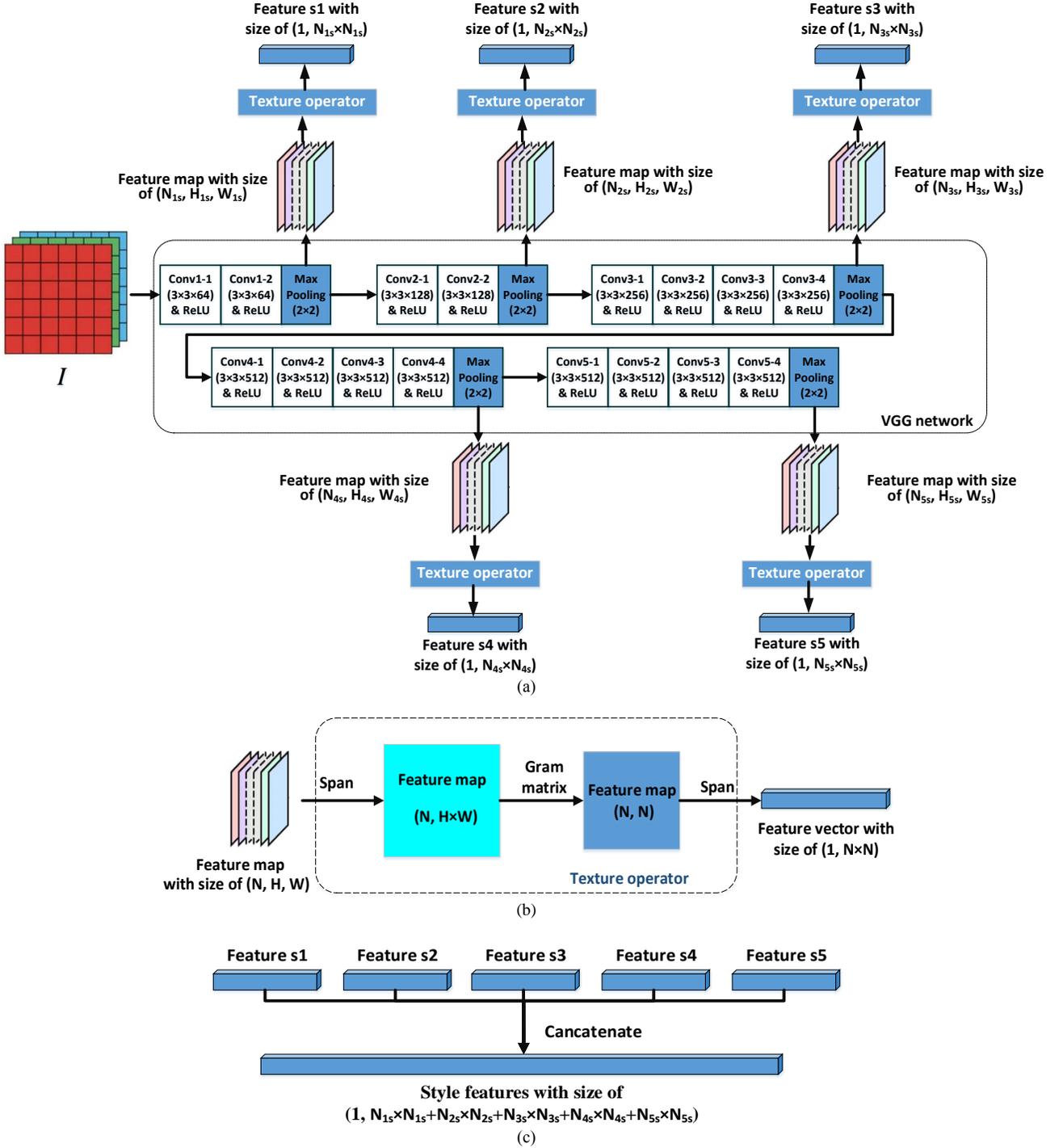

Fig. 2. Flowchart of the extraction of the style features. (a) Extraction of each part of the style features; (b) Texture operator; (c) Feature concatenation.

In each iteration, $\mathbf{W}_i^k$ fluctuates around $\mathbf{W}_i^0$, $i = 1,2,\dots,16$, with the normal distribution. Therefore, based on the extraction framework shown in Figs. 1 and 2, it is ensured that the extracted features based on $\mathbf{W}_i^k$, i.e., $C^k(\cdot)$ and $S^k(\cdot)$ $k \geq 1$, can also effectively extract the feature subspaces of the semantic content and the style with similar properties, respectively. Furthermore, by randomization in (6), $C^k(\cdot)$, $k \geq 0$, are different with each other in each iteration, which also holds true for $S^k(\cdot)$, $k \geq 0$.

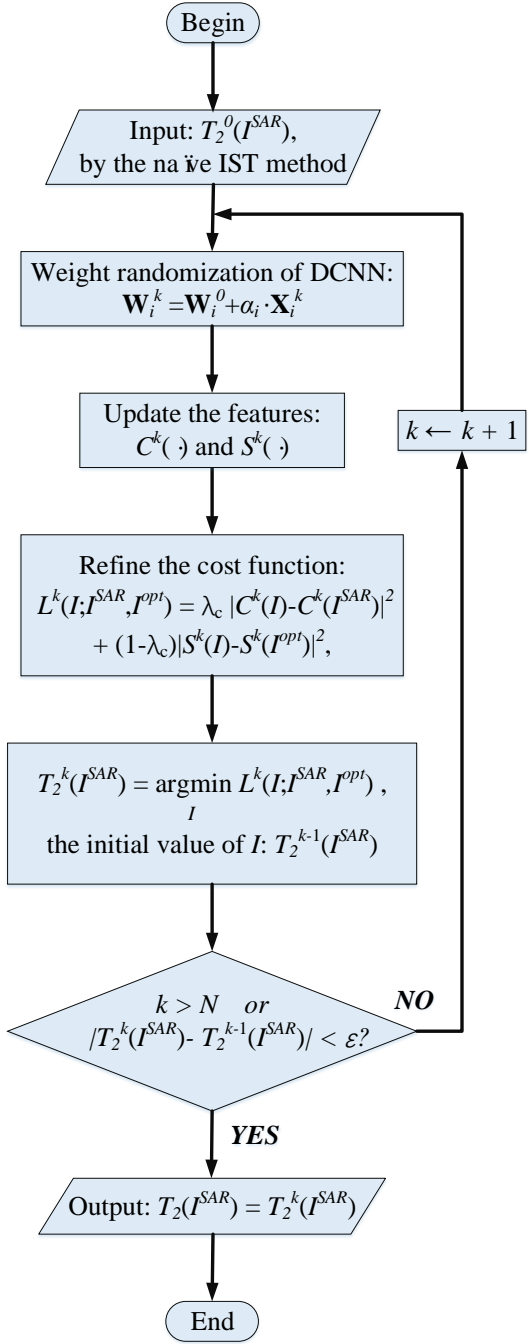

Fig. 3. Flowchart of the IIST strategy. Here $N$ is the max number of iteration and $\varepsilon$ is the convergence threshold of the iteration.

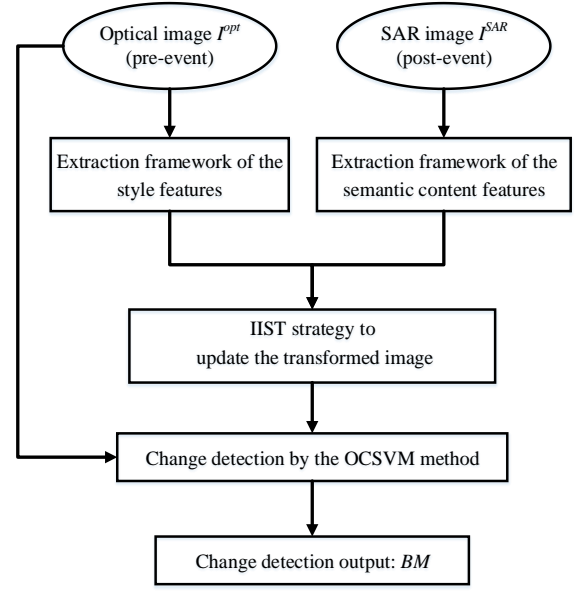

Fig. 4. Flowchart of the proposed method of change detection via deep homogeneous feature fusion (DHFF).

the semantic content of $I^{SAR}$ and the styles of $I^{opt}$, $T_2^k(I^{SAR})$ can be infinitely close to the real post-event optical image.

The naïve IST method uses the pre-trained VGG network with the fixed filter weights for style transferring. In other words, $T_2^0(I^{SAR})$, derived by minimizing the single cost function $L^0(\cdot)$ with $C^0(\cdot)$ and $S^0(\cdot)$, is the result of the naïve IST method, which means the feature homogeneity is only achieved in a single feature subspace with limited semantic content features $C^0(\cdot)$ and style features $S^0(\cdot)$. Compared with the naïve IST method, the proposed method achieves the

---

**Algorithm 1**. IIST Strategy with the VGG Network of Randomized Filter Weights

---

**Input:**

    Pre-event optical image: $I^{opt}$

    Post-event SAR image: $I^{SAR}$

**Output:**

    The transformed image: $T_2(I^{SAR})$ that achieves feature homogeneity for change detection on homogeneous images

**Algorithm procedure:**

    1. Building the extraction framework of the semantic content and the style features:

        a) Build the extraction framework of the semantic content features according to Fig. 1.

        b) Build the extraction framework of the style features according to Fig. 2.

    2. Iterative strategy:

        a) For $k = 0$, use $I^{SAR}$ as the initial image to derive $T_2^0(I^{SAR})$, according to (4) and (5).

        b) For $k \geq 1$, use $T_2^{k-1}(I^{SAR})$ as the initial image to derive $T_2^k(I^{SAR})$, according to (4), (5), and (6).

        c) Stop criterion: $|T_2^{k+1}(I^{SAR}) - T_2^k(I^{SAR})| < \varepsilon$ $or$ $k > N$

---

Intuitively, the IIST strategy in (4) and (5) is expected to converge. For a given image, $C^k(\cdot)$ and $S^k(\cdot)$, $k \geq 0$, are the feature subspace that describes the semantic content and styles, respectively. After a number of iterations, $C^k(\cdot)$ and $S^k(\cdot)$, $\forall k \geq 0$, extracted by the DCNN with randomized weights, are expected to cover the whole feature space of the image. At this time, $T_2^k(I^{SAR})$ will converge because the feature homogeneity has been already achieved in the semantic content and the style described by these feature subspaces. Ideally, when all of $C^k(I^{SAR})$ and $S^k(I^{opt})$, $k \geq 0$, completely cover feature homogeneity in multiple feature subspaces described by $C^0(\cdot), C^1(\cdot), C^2(\cdot),\cdots$ and $S^0(\cdot), S^1(\cdot), S^2(\cdot),\cdots$. These new feature subspaces enhance the description ability of the semantic content and the style, greatly. Therefore, the updated transformed image in (4) promotes the semantic content feature homogeneity with $I^{SAR}$ and the style feature homogeneity with $I^{opt}$. When the iterations end, the feature homogeneity of the transformed image will be maximized. The IIST strategy that achieves the feature homogeneity in the transformed image is shown in Algorithm 1 and Fig. 3, where $N$ is the max number of iterations and $\varepsilon$ is the convergence threshold.

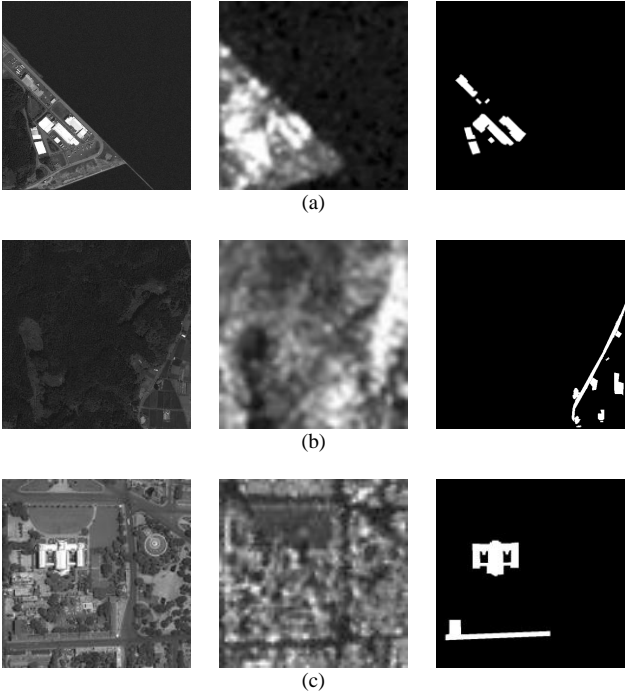

(a)

(b)

(c)

Fig. 5. Coregistered optical and SAR images of the experimental datasets with the ground truths (left: optical; middle: SAR; right: ground truth): (a) The first dataset; (b) The second dataset; (c) The third dataset.

The change detection result $BM$ is derived based on $T_2(I^{SAR})$, according to (1). The commonly used change detection method OCSVM [35] for optical images is applied on $T_2(I^{SAR})$ and $I^{opt}$, both of which are in the optical feature space, to derive $BM$.

In summary, the flowchart of the proposed DHFF method is shown in Fig. 4.

## IV. EXPERIMENTAL RESULTS

In this section, the 2011 Tōhoku earthquake (on March 11, 2011, with $M_w$ 9.0 measured on Richter Scale) and the Haiti earthquake (on January 12, 2010, with $M_w$ 7.0 measured on Richter Scale) are used as the study cases. Three real datasets, all of which consist of a pre-event optical image and a post-event SAR image, are used to evaluate the performance of the proposed method. The information of the datasets is summarized in Table I. For the first two datasets, the SAR images were collected by the RADARSAT-2 satellite and the optical images were obtained by the GeoEye-1 satellite. For the third dataset, the SAR and optical images are acquired by

COSMO-SkyMed and Quickbird satellites, respectively. As shown in Fig. 5, in the experimental datasets, the quality of the optical images is better than that of the SAR images with much higher resolutions and more details of the ground objects. Therefore, we select the optical image as the target feature space for homogeneous transformation, to reduce the loss of semantic content during image transformation. The ground truths of changed regions are provided by Yanagawa (the first and the second datasets) [36] and United Nations Institute for Training and Research (the third dataset) [37] To deal with different resolutions between the heterogeneous images, we use the bilinear interpolation [38] to equalize their resolutions for the homogeneous transformation. Besides, in the experimental datasets, both of the SAR and the optical images are coregistered by visual selection of the controlling points [39].

TABLE I
INFORMATION OF THE EXPERIMENTAL DATASETS

| Sensor Type | First and Second dataset | | Third dataset | |
| --- | --- | --- | --- | --- |
| | Optical | SAR | Optical | SAR |
| Satellite | GeoEye-1 | RADARSAT-2 | Quickbird | COSMO-SkyMed |
| Acquisition Date | Sept. 29, 2009 (Pre-) | Mar. 12, 2011 (Post-) | Jul. 27, 2009 (Pre-) | Jan. 21, 2010 (Post-) |
| Resolution | 0.41 m | 8 m | 0.6 m | 6 m |

The first dataset corresponds to a coastal area in Rikuzentakata, as shown in Fig. 5(a). The buildings near the coasts were severely damaged by the earthquake and tsunami [36]. The pre-event optical image was acquired with the size of 1250×1250 pixels in September 2009. The post-event SAR image was with the size of 64×64 pixels, collected in March 2011. The second dataset corresponds to a suburban area of Iwate prefecture, as shown in Fig. 5(b), which was also damaged seriously after the earthquake. The SAR image is with the size of 105×105 pixels and the size of the optical image is 2048×2048 pixels. The third dataset, collected by another group of SAR and optical satellites, is shown in Fig. 5(c). The dataset focuses on an urban area of Port-au-Prince, destroyed seriously by the Haiti earthquake. The sizes of SAR and optical images are 64×64 pixels and 640×640 pixels, respectively.

In the experiments, we compare the proposed method with the following methods: linear regression [19] (denoted by LR), SCCN [7], and HPT [20] in terms of the performance of change detection for heterogeneous images. The OCSVM method directly applied on the original SAR and optical images (denoted by OCSVM_O) is also included in the comparison to better validate the effects of the IIST strategy in the proposed method. Among these methods, LR is the basic model for homogeneous transformation. The other two are the state-of-the-art methods. We also test a method of only using the image $T_2^0(I^{SAR})$ for change detection, which is named HFF, to illustrate the separate effect of the proposed IIST strategy. The HFF method can be seen as the direct application of the naïve IIST method without any improvement. Compared with the proposed DHFF method, the HFF method validates the effectiveness of the procedure of segregated extraction of the semantic content and the style features. In the two state-of-the-art methods for comparison, HPT [20] uses pixel values as the transformation features, while SCCN [7] builds a convolutional neural network with 4 layers for each heterogeneous image to extract features. As can be seen, both of the pixel values and the output of the SCCN are not deep/abstract enough to extract the high-level semantic content. Therefore, the semantic content may be susceptible to corruption in the homogeneous transformation process, especially in the regions with multiple ground objects and rich semantic content, leading to inaccurate change detection results.

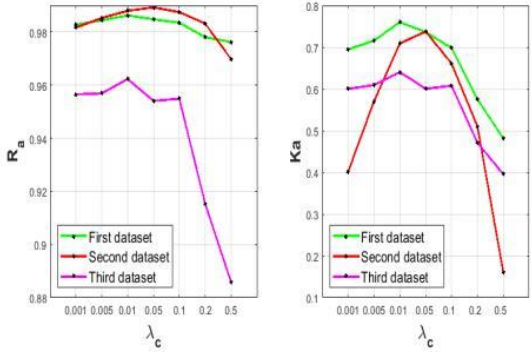

Fig. 6. Relationship between the overall detection performance $R_a/Ka$ and $\lambda_c$ on the different datasets.

The quantitative evaluations of the above six methods are carried out based on the following criteria [40] with four frequently used measurements, as given below:

$$R_a = \frac{m_a + m_c}{M}, \quad R_p = \frac{m_a}{M_d}, \quad R_r = \frac{m_a}{M_c}, \quad Ka = \frac{R_a - p_e}{1 - p_e}, \quad (7)$$

where $R_a$, $R_p$, $R_r$, and $Ka$ are the accuracy rate, the precision rate, the recall rate, and the Kappa index, respectively; $m_a$ and $m_c$ are the numbers of changed and unchanged pixels which are correctly detected, respectively; $M_d$ is the total number of pixels detected as change by the method; $M_c$ is the total number of truly changed pixels; $M$ is the number of all the pixels in the image; the Kappa index, $Ka$, is commonly used to evaluate the detection quality comprehensively with $P_e$ as the hypothetical probability of random agreements[41]. Among the four measurements, $R_a$ and $Ka$ evaluate the overall performance of detection.

In the following, we first discuss the influence of the related parameters on the performance of the proposed DHFF method, i.e., $\lambda_c$, $\{\alpha_i, i = 1,2,3, \dots, 16\}$, $N$, and $\varepsilon$. Then, the proposed method is compared with several change detection methods on the three real datasets.

### A. Parameters Setting

1) *Effect of the Parameter* $\lambda_c$: In the proposed method, the value of $\lambda_c \in (0,1)$ in (3) is related to the influence of the semantic content and the style features on the homogeneous transformation. A too small $\lambda_c$ means little consideration for the semantic content features, leading to less preservation of the image semantic content in the transformation. If $\lambda_c$ is too large, the style features will be underestimated, resulting in insufficient transformation of the feature space. As the dimensions of semantic content features are much greater than those of style features (dimensions are largely reduced by the Gram matrix), the naïve IST method assigns small values of $\lambda_c$ to balance the influence of semantic content and styles.

Similarly, we set $\lambda_c \in \{0.001, 0.005, 0.01, 0.05, 0.1, 0.2, 0.5\}$, which distributes dense around 0, to evaluate its relationship with the overall detection performance, $R_a$ and $Ka$. The values of $R_a$ and $Ka$ versus $\lambda_c$ are shown in Fig. 6. In Fig. 6, the detection performance is satisfactory for all of the three datasets when $\lambda_c \in [0.01, 0.05]$. Specifically, we choose the value of $\lambda_c$ to be 0.01, 0.05, and 0.01 for the first, the second, and the third datasets in the experiments, respectively.

2) *Effects of the Parameters* $\{\alpha_i, i = 1,2,3, \dots, 16\}$ : $\{\alpha_i, i = 1,2,3, \dots, 16\}$ control the intensity of the noise added to the pre-trained filter weights of DCNN. In the proposed method, a too large $\alpha_i$ makes the DCNN deviated far from the fine-tuned VGG network and thus weaken the ability of the additional new feature subspaces for homogeneous transformation. For a small $\alpha_i$, the ability of the additional new feature subspaces is limited. Here we set all the $\alpha_i$ to be 1 as an empirical and compromised selection [42].

3) *Effects of the Parameters* $N$ *and* $\varepsilon$: In the experiments, $N$ and $\varepsilon$ are used as the thresholds to control the speed of the IIST strategy. A too large $N$ keeps iterating until $I_{itr}$ converges, leading to time wasting. If $N$ is too small, the iteration will be ended early before it converges. In the experiments, $N = 100$ is suggested as a satisfactory setting to guarantee the iteration convergence. The value of $\varepsilon$ should be small enough to keep the stability of the convergence. Here $\varepsilon$ is set to be 0.01 as a relatively weak constraint.

### B. Results on the First Dataset

The experimental results corresponding to the first dataset are shown in Figs. 7 and 8. Fig. 7 shows the transformed images: the initial transformed image $T_2^0(I^{SAR})$, the intermediate image $T_2^5(I^{SAR})$ in the iteration process, and the final transformed image $T_2(I^{SAR})$ generated after the iteration ends. Fig. 8 compares the detection results of different methods.

Fig. 7 presents the transformed images derived by the proposed IIST strategy. In Fig. 7(a), i.e., $T_2^0(I^{SAR})$, derived by the naïve IST method, the style is changed and the semantic content is still preserved, validating the effectiveness of the separate extraction of the semantic content and the style features. Because of the extensive information carried in the deep-level features and the significant difference between the style of optical and SAR images, the DCNN with limited filter weights cannot achieve the feature homogeneity, resulting in vague contours of ground objects and massive bright inhomogeneous regions in the water area, as shown in Fig. 7(a). As a result, the feature homogeneity is not achieved, compared with the real optical image $I^{opt}$, as shown in Fig. 7(d). In Fig. 7(b), after 5 iterations, the contours of the ground objects become clear and the bright inhomogeneous regions are reduced sharply. Compared with Fig. 7(a), Fig. 7(b) is much more homogeneous with the optical image, validating the effectiveness of the feature subspaces added in each loop of the iterations. Fig. 7(c) shows the final transformed image, $T_2(I^{SAR})$, of which the feature homogeneity is finally achieved with the optical image. Compared with Fig. 7(a) and (b), Fig. 7(c) eliminates most of the bright inhomogeneous regions in the upper- and the lower-right parts of the images. The edges of the lands and the buildings are much clearer than those of Fig. 7(b). Besides, the narrow breakwater in the lower-right is also preserved. Therefore, it is necessary to utilize the proposed IIST strategy that includes multiple feature subspaces extracted by the DCNN with randomized filter weights, to generate the transformed image with feature homogeneity.

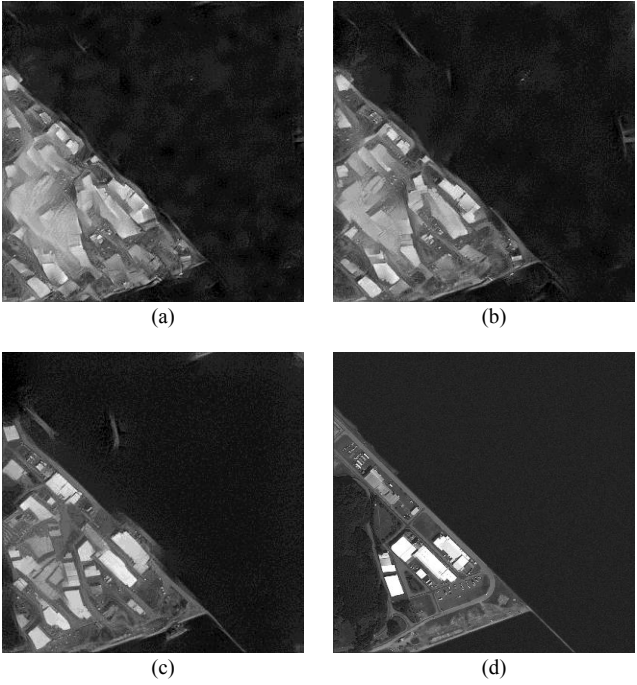

(a)        (b)

(c)        (d)

Fig. 7. Transformed optical images of the first dataset. (a) $T_2^0(I^{SAR})$: the initial transformed image derived by the naïve IST method, i.e., $k$=0; (b) $T_2^5(I^{SAR})$: the intermediate result of the proposed IIST strategy after 5 iterations, i.e., $k$=5; (c) $T_2(I^{SAR})$: the final transformed image derived by the proposed IIST strategy, in this case, $k$=65 when the iteration ends; (d) $I^{opt}$: the real pre-event optical image for comparison. With the increase of the iteration loops, the feature space of the transformed image is more homogeneous with that of the optical image, shown in (d).

The proposed DHFF method is compared with other methods in Fig. 8. In Fig. 8(a), the change detection based on linear regression causes massive false alarms. The performance of change detection is unsatisfied caused by the limited properties of the features of linear regression. As can be seen in Fig. 8(a), most of the buildings, roads, and coasts are not sufficiently transformed and thus detected as false alarms. Different from Fig. 8(a), Fig. 8(b) and (c) eliminate most of the false alarms in the buildings, roads, and coasts because both of the SCCN and the HPT methods extract more sophisticate features for homogeneous transformation by transfer learning and neural networks, respectively. However, there still exist considerable false alarms and missed targets in the lower-left part of the results. The corruption of the image semantic content in the homogeneous transformation is the main reason. As can be seen in Fig. 5(a), the lower-left part of the optical image includes various kinds of ground objects, e.g., multiple buildings, roads, and forests, with richer semantic content than other parts of the image. If represented by the low-level features, the semantic content of these regions is more difficult to be preserved in the homogeneous transformation than other regions. By the proposed separate extraction of the semantic content and the style features based on the high-level features with DCNN, most of the false alarms and missed targets are removed, as shown in Fig. 8(d) and (f). In Fig. 8(e) and Table II, the detection performance of the OCSVM_O method is poor, illustrating the infeasibility of direct employment of the OCSVM method on the original SAR and optical images.

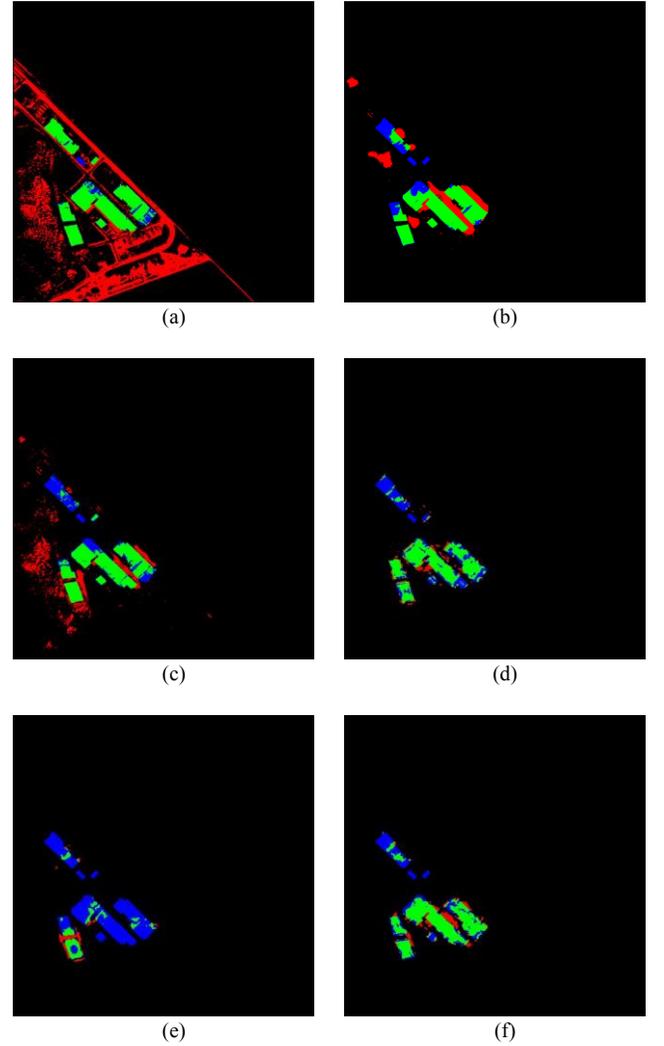

(a)        (b)

(c)        (d)

(e)        (f)

Fig. 8. Comparisons of the change detection results/error maps based on the first dataset. Here green color indicates the correct detection of regions of change, red color implies the regions of false alarms, blue color represents the areas of missed targets, and black color illustrates the regions of unchange that are correctly detected. The error maps are achieved by (a) LR; (b) SCCN; (c) HPT; (d) HFF; (e) OCSVM_O; (f) DHFF. As can be seen, the proposed DHFF method (f) achieves the best performance.

Compared with Fig. 8(d), Fig. 8(f), derived by the proposed DHFF method, detects the regions of change that are more complete with less missed targets. It is because of the feature homogeneity achieved by the IIST strategy. It ensures the homogeneous feature space for the subsequent change detection method. By employing the proposed separation of the semantic content and the style features with iterative minimization based on the VGG network with randomized filter weights, the regions of change are well detected and most of the missed targets and false alarms are eliminated, validating the effectiveness of the proposed method.

Apart from the visual comparisons, the results of the above methods are also compared in terms of quantitative evaluations.

The values of the accuracy rate $R_a$, the precision rate $R_p$, the recall rate $R_r$, and the Kappa index $Ka$, produced by these methods are listed in Table II. Compared with four other methods, both the HFF method and the proposed new DHFF method performing much better on $R_a$, $R_p$, and $Ka$ because of separate extraction of the semantic content and the style features. Although the LR method achieves the highest recall rate, it produces the lowest precision rate induced by the limited transformation ability of the features of linear regression, leading to the unsatisfactory $R_a$ and $Ka$. By the proposed IIST strategy, the DHFF model achieves the overall detection performance ($R_a / Ka$) better than the HFF method based on the VGG network with limited filter weights. Here the precision rate $R_p$ of the HFF method is a little higher than that of the proposed DHFF method. The reason is that the feature spaces of the heterogeneous optical and SAR images in most regions of change are similar. As shown in Fig. 5(a), most regions of change are covered with the same bright intensity in both SAR and optical images. The similarity makes the feature space of these regions easy to be transformed. Therefore, the naïve IST method can manage the homogeneity of large parts of these regions, leading to the detection of these regions with higher precision rate $R_p$. However, the edges of these regions are more difficult to be transformed because their feature spaces are much more different. Therefore, the naïve IST method fails in the transformation of the edges, resulting in lower recall rate $R_r$, as shown in Table II. By applying the IIST strategy in the proposed DHFF method, most edges of these regions are well-transformed and detected, as shown in Fig. 8(f). The recall rate $R_r$ is thus improved with the overall performance $R_a / Ka$.

### TABLE II
COMPARISONS OF DETECTION METHODS BASE ON THE FIRST DATASET (%)
(The **boldface** indicates the best results.)

| Method | $R_a$ | $R_p$ | $R_r$ | $Ka$ |
|---|---|---|---|---|
| LR | 91.55 | 26.21 | **88.21** | 37.34 |
| SCCN | 97.77 | 63.46 | 73.35 | 66.90 |
| HPT | 97.46 | 59.04 | 69.68 | 62.61 |
| HFF | 98.40 | **84.93** | 61.43 | 70.49 |
| OCSVM_O | 97.07 | 68.55 | 17.08 | 26.40 |
| **DHFF** | **98.63** | 84.66 | 70.11 | **76.00** |

### C. Results on the Second Dataset

Different from the first dataset, the second dataset is covered with more complicated backgrounds due to the dense forests with the complex style, increasing the difficulty for change detection.

Same as that in the first experiment, the transformed images are shown in Fig. 9(a). In Fig. 9(a), i.e., $T_2^0(I^{SAR})$, many forest regions are covered with bright intensity, indicating the inhomogeneity with the optical image. Compared with Fig. 9(a), Fig. 9(b) is more homogeneous with the optical image, demonstrating the effectiveness of the additional new feature subspaces extracted by the DCNN with randomized filter weights. However, Fig. 9(b) does still not achieve the feature homogeneity as its textures of the farmland shown in the lower-right of the image are largely different from those of the optical image. Compared with Fig. 9(a) and (b), Fig. 9(c) is the more homogeneous. It illustrates the effectiveness of the proposed IIST strategy with randomized filter weights.

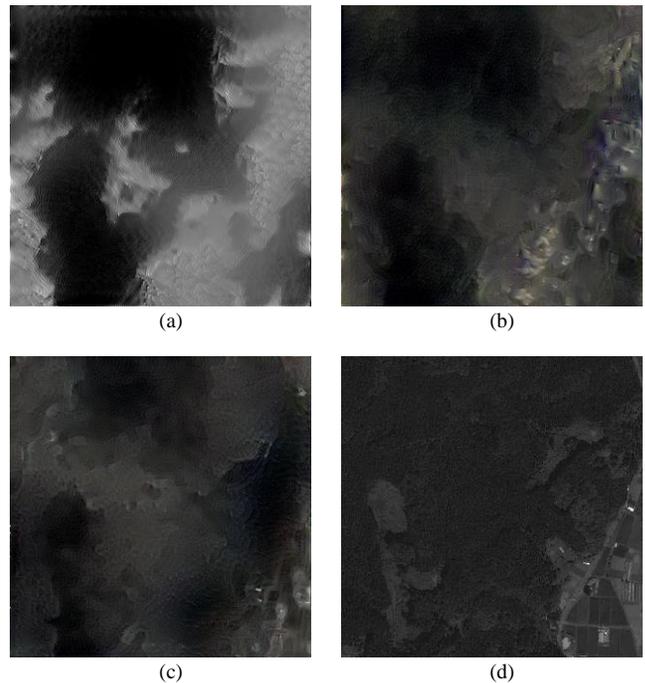

Fig. 9. Transformed optical images of the second dataset. (a) $T_2^0(I^{SAR})$: the initial transformed image derived by the naïve IST method, i.e., $k=0$; (b) $T_2^5(I^{SAR})$: the intermediate result of the proposed IIST strategy after 5 iterations, i.e. $k=5$; (c) $T_2(I^{SAR})$: the final transformed image derived by the proposed IIST strategy, in this case, $k=93$ when the iteration ends; (d) $I^{opt}$: the real pre-event optical image for comparison. Same as that in Fig. 7, with the increase of the iterations, the feature homogeneity is improved in each loop of the iteration, and finally achieved by the proposed IIST strategy.

The comparisons between the proposed method and other methods are demonstrated in Fig. 10. The linear regression can hardly describe the transformation relationship between the heterogeneous optical and SAR images and is only capable of describing and transforming simple ground objects, leading to massive false alarms in the non-forest regions shown in Fig. 10(a). As shown in Fig. 10(b) and (c), this situation is improved by applying neural networks (SCCN) and transfer learning (HPT) for homogeneous transformation. However, both of the two methods still cause reasonable false alarms and missed targets in the lower-right parts of the results because of corrupted semantic content in homogeneous transformation. In Fig. 10(e) and Table III, similar to the first experiment, the detection performance of the OCSVM_O method is unsatisfactory. It illustrates that the limited effectiveness of the supervised OCSVM method and validates the effectiveness of homogeneous transformation with the IIST strategy. Compared with the above methods, the HFF and the proposed DHFF methods based on the segregation of the semantic content and the style avoid the corruption of the image content in homogeneous transformation, especially in the regions with various ground objects and rich semantic content. In Fig. 10(d) and (f), most of the false alarms and the missed targets in Fig. 10(a), (b), (c), and (d) are eliminated.

Compared with Fig. 10(d), Fig. 10(f) shows the results with much less false alarms and missed targets. As the proposed DHFF method uses he IIST strategy to update the transformed image iteratively, the homogeneity of the semantic content and the style features is achieved finally, leading to improvement of detection performance.

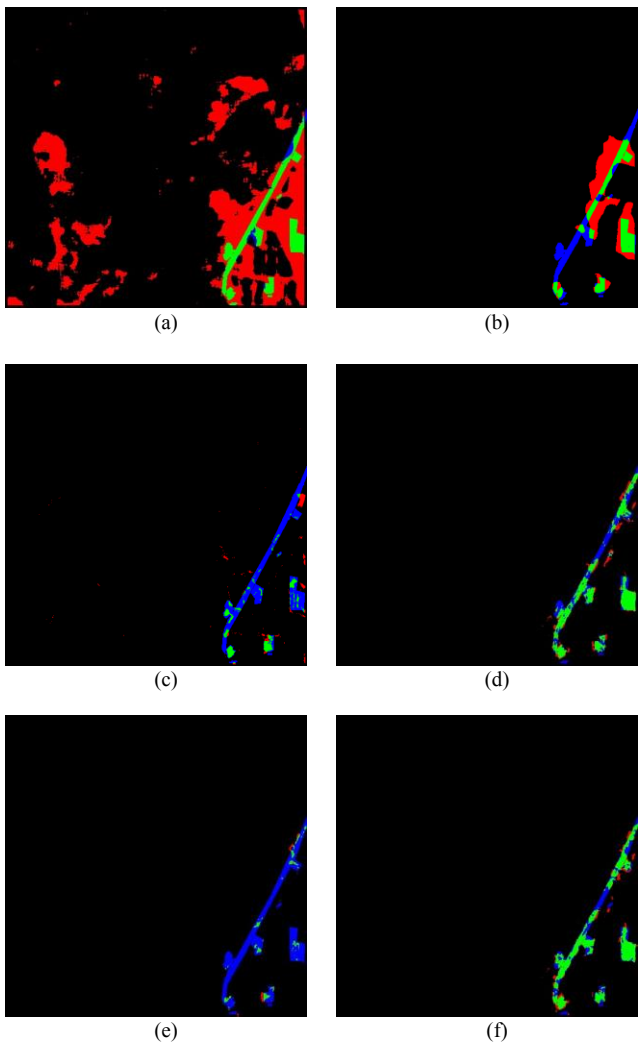

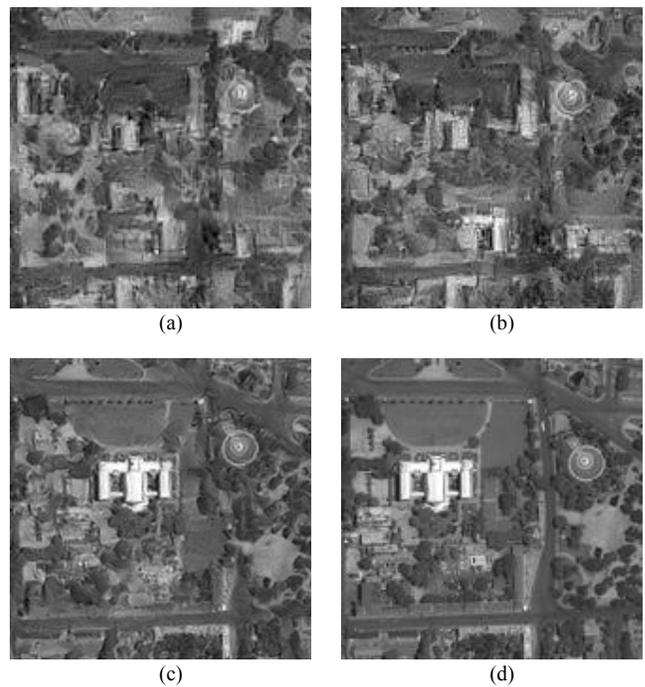

Fig. 11. Transformed optical images of the third dataset. (a) $T_2^0(I^{SAR})$: the initial transformed image derived by the naïve IST method, i.e., $k$=0; (b) $T_2^5(I^{SAR})$: the intermediate result of the proposed IIST strategy after 5 iterations, i.e., $k$=5; (c) $T_2(I^{SAR})$: the final transformed image derived by the proposed IIST strategy, in this case, $k$=72 when the iteration ends; (d) $I^{opt}$: the real pre-event optical image for comparison. With the increase of the iteration loops, the feature space of the transformed image is more homogeneous with that of the optical image, shown in (d).

Fig. 10. Comparisons of the change detection results/error maps based on the second dataset. Here green color indicates the correct detection of regions of change, red color implies the regions of false alarms, blue color represents the areas of missed targets, and black color illustrates the regions of unchange that are correctly detected. The error maps are achieved by (a) LR; (b) SCCN; (c) HPT; (d) HFF; (e) OCSVM_O; (f) DHFF. Similar to Fig. 8, the proposed DHFF method (f) performs the best.

The quantitative evaluations of these six methods are also compared in Table III, including the accuracy rate $R_a$, the precision rate $R_p$, the recall rate $R_r$, and the Kappa index $Ka$. We can see that the propose DHFF method achieves much higher $R_a/R_p/Ka$ than the other five methods. Although the LR method produces highest $R_r$, it causes the lowest $R_p$ and the worst overall performance $R_p$ and $Ka$. It is because that the linear regression cannot sufficiently transform most of the ground objects, e.g., roads, buildings, and farmland, leading to massive false alarms. The performance of the SCCN and the HPT methods is unsatisfactory. It is because that they corrupt the semantic content in homogenous transformation. Besides, the HFF method produces the second best $R_a/R_p/Ka$ only to the proposed DHFF method. This demonstrates the effectiveness of the semantic content and the style features separately extracted by the IST with DCNN. Besides, it also illustrates the sufficiency and accuracy of the transformed image derived by the proposed IIST strategy based on the VGG network with the randomized filter weights.

TABLE III
COMPARISONS OF DETECTION METHODS BASE ON THE SECOND DATASET (%)
(The **boldface** indicates the best results.)

| Method | $R_a$ | $R_p$ | $R_r$ | $Ka$ |
|---|---|---|---|---|
| LR | 87.02 | 14.03 | **90.52** | 7.48 |
| SCCN | 96.79 | 36.76 | 54.79 | 42.42 |
| HPT | 97.92 | 65.98 | 20.08 | 30.03 |
| HFF | 98.79 | 80.59 | 62.49 | 69.78 |
| OCSVM_O | 98.03 | 77.17 | 9.00 | 15.71 |
| **DHFF** | **98.93** | **83.31** | 67.03 | **73.77** |

### D. Results on the Third Dataset

The experimental results of the third dataset are shown in Figs. 11 and 12. Different from the first two datasets, the third dataset consists of the optical and the SAR images collected by another group of satellites. Fig. 11 shows the transformed images and Fig. 12 compares the proposed DHFF method with other methods.

The transformed images are shown in Fig. 11. As shown in Fig. 11(a), the initial image $T_2^0(I^{SAR})$ is not homogeneous with the optical image with vague contours of the ground objects. Compared with Fig. 11(a), Fig. 11(b), the intermediate results after 5 iterations, is more homogeneous in the ground objects with more distinct contours. In Fig. 11(c), after the iteration ends, the feature space of the transformed image is homogeneous with that of the optical image, as shown in Fig. 11(d). It validates the effectiveness of the IIST strategy.

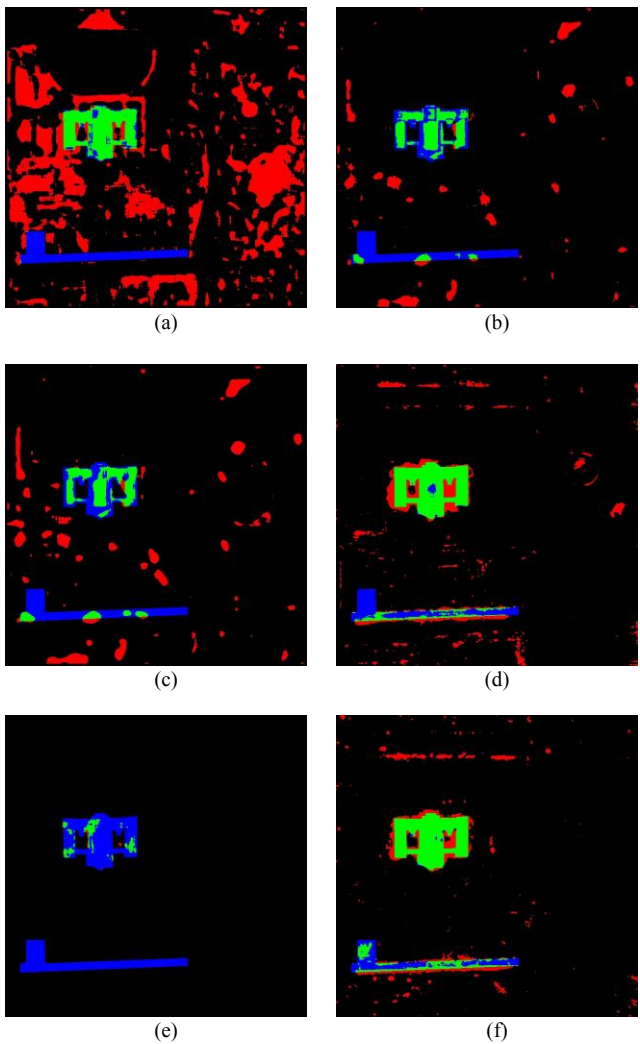

Fig. 12. Comparisons of the change detection results/error maps based on the third dataset. Here green color indicates the correct detection of regions of change, red color implies the regions of false alarms, blue color represents the areas of missed targets, and black color illustrates the regions of unchange that are correctly detected. The error maps are achieved by (a) LR; (b) SCCN; (c) HPT; (d) HFF; (e) OCSVM_O; (f) DHFF. Similar to Figs. 8 and 10, the proposed DHFF method (f) performs the best.

The comparisons of the proposed and other methods are shown in Fig. 12. Similar to the first two experiments, the performance of the LR method in Fig. 12(a), is unsatisfactory because of the limited transformation ability of linear regression. The SCCN and the HPT methods perform better than the LR method, as shown in Fig. 12(b) and (c). This illustrates the effectiveness of the homogeneous transformation based on neural networks and transfer learning. Similar to the previous experiments, the detection performance of the OCSVM_O method in Fig. 12(e) is poor, showing that the direct utilization of the OCSVM method on the SAR and optical images is infeasible. By separate extraction of semantic content and styles of the images, the HFF and the DHFF methods achieve better results, as shown in Fig. 12(d) and (f).

The feature space of the transformed image, derived by the HFF method, is still not homogeneous with that of the optical image. This leads to the false alarms in the inhomogeneous regions, as shown in Fig. 12(d). With the proposed IIST strategy, the feature homogeneity is improved in the final transformed image, leading to the elimination of most false alarms in Fig. 12(f).

The quantitative evaluations of the above methods are also compared in Table IV, including the accuracy rate $R_a$, the precision rate $R_p$, the recall rate $R_r$, and the Kappa index $Ka$. Although the OCSVM_O method achieves the highest $R_p$, it produces the lowest $R_r$ and the second lowest $Ka$. The OCSVM method can hardly learn the massive and complicate change patterns directly from the heterogeneous optical and SAR images, leading to detection of a few regions of change. We can see that the HFF and the propose DHFF methods achieve better performance than the other methods. This demonstrates the importance of separation of the semantic content and the style features in homogeneous transformation. Besides, the proposed DHFF method achieves better quantitative detection performance than the HFF method.

TABLE IV
COMPARISONS OF DETECTION METHODS BASE ON THE THIRD DATASET (%)
(The **boldface** indicates the best results.)

| Method | $R_a$ | $R_p$ | $R_r$ | $Ka$ |
|---|---|---|---|---|
| LR | 79.30 | 12.61 | 52.78 | 13.35 |
| SCCN | 94.64 | 45.84 | 38.67 | 39.16 |
| HPT | 93.65 | 37.22 | 38.83 | 34.67 |
| HFF | 95.25 | 51.62 | 70.07 | 57.02 |
| OCSVM_O | 95.38 | **92.74** | 8.54 | 14.92 |
| **DHFF** | **98.23** | 58.19 | **76.23** | **64.04** |

### E. Analysis of the Time Consumption

The time of performing the DHFF method consists of two parts: (1) homogeneous transformation and (2) training and inferencing the OCSVM classifier. As mentioned above, the homogeneous transformation converges in limited iterations. The training and inferencing time of the OCSVM classifier, as a type of SVM, is limited. Therefore, the consumption of the DHFF method is controllable.

The computational time of different methods based on the homogeneous transformation is shown in Table V. The hardware platform is a server with an Intel(R) Core(TM) i9-7980XE CPU, 128GB RAM, and a NVIDIA Titan RTX Graphics card inside. The software platform is MATLAB 2018b, Python3.5, and TensorFlow 1.14 with the operation system Ubuntu 16.04. The running time is measured only on the first dataset by over 20 trials as the task is the same with that of the other two datasets. As there exist supervised and unsupervised methods for comparison, we put the training and inferencing time together that is convenient to compare with the other supervised methods. The time consumption of the LR method is the least because of the simple linear transformation, but it performs the worst as shown in Tables II, III, and IV. The proposed DHFF method costs the longest time in total because it performs the iterative update of the transformed image.

| Method | transformation | training & inferencing | total |
|--------|----------------|------------------------|-------|
| LR | 0.162s | 12.461s | 12.623s |
| SCCN | 1780.146s | 4115.089s | 5895.235s |
| HPT | 1042.055s | 254.091s | 1296.146s |
| HFF | 74.946s | 12.385s | 87.331s |
| DHFF | 9104.122s | 11.960s | 9116.082s |

## V. CONCLUSIONS

In this paper, we present a new method, namely DHFF, for change detection in heterogeneous optical and SAR images via deep homogeneous feature fusion. Different from the existing method based on the homogeneous transformation, the proposed method can transform the heterogeneous images into the same feature space accurately, leading to better performance of change detection at the cost of the increased computational complexity. By the image style transfer (IST), which is originally used to render a natural image into specific artistic styles, the new DHFF method separately extracts the semantic content and the style features based on different layers of DCNN, avoiding the corruption of the image semantic content in the homogeneous transformation.

Furthermore, to achieve the feature homogeneity for change detection, a new iterative IST (IIST) strategy is proposed. Different from the naïve IST method that uses a single cost function based on the feature subspace with limited style features for style transferring, the proposed method minimizes the cost function in each iteration that measures the feature homogeneity in additional new feature subspace, to update the transformed image with promotion of the feature homogeneity. Therefore, the requirements for change detection in homogeneous optical images are met after the iteration converges.

In the proposed DHFF method, different layers of the DCNN are used as the extraction framework to separate the semantic content and the style features, avoiding the corruption of the semantic content in the homogeneous transformation. Then, the filter weights of the DCNN in the above extraction framework are randomized to generate additional new feature subspaces. These feature subspaces are utilized to build multiple cost functions to improve the feature homogeneity of the transformed image with the IIST. Finally, a commonly used change detection method for optical images is applied on the pre-event optical image and the transformed post-event image to generate the final detection results. The proposed method preserves the semantic content in the homogeneous transformation by the deep-level features from the DCNN, especially in the regions that are vulnerable to corruption with multiple ground objects and rich semantic content.

Experiments are conducted on three real remote sensing datasets. Compared with the existing methods based on the homogeneous transformation, the proposed DHFF method avoids the corruptions of semantic content in the transformed images and improves the feature homogeneity by the IIST strategy, leading to accurate detection of the changed regions with multiple ground objects and complex scenes. The quantitative evaluations demonstrate the superior performance of the proposed method in terms of accuracy rate and Kappa index, especially in the regions with rich semantic content.

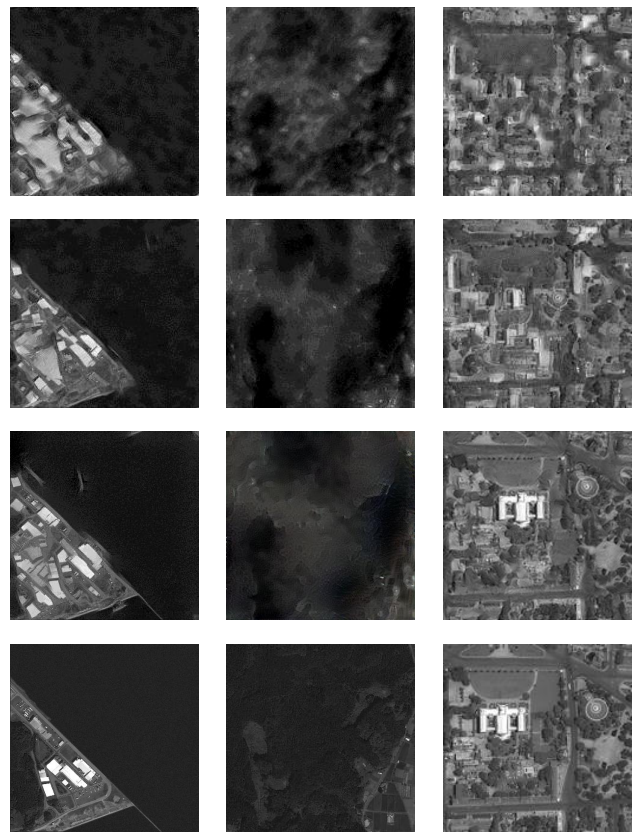

Fig. 13. Transformed optical images based on the semantic content features with different layer hyperparameters. As for the images above, different columns correspond to different datasets. (The first, the second, and the third columns are from the first, the second, and the third datasets, respectively.) Different rows represent different layer hyperparameters of the semantic content. (The first, the second, and the third rows represent the images with the semantic content from Conv3-4, Conv4-4, and Conv5-4, respectively. The fourth row gives the real optical images for comparison.) As can be seen, the images transformed by using Conv5-4 as the layer hyperparameter are the most homogeneous with the optical images.

## APPENDIX I

Different layer hyperparameters are compared to select the optimal semantic content features. The deepest convolutional layers in the third, the fourth, and the fifth scales of the VGG network, i.e., Conv3-4, Conv4-4, and Conv5-4, are chosen for comparison. We compare the convolution layers because they keep semantic content without non-linear operations. The layers in the first and the second scales are not considered as they are not deep enough for image style transfer [28].

Fig. 13 compares the transformed images derived by using different layer hyperparameters for extracting semantic content features. As can be seen in Fig. 13, Conv5-4 (in the third row) achieves the most homogeneous results. It validates Conv5-4 as the suitable layer hyperparameter to extract the semantic

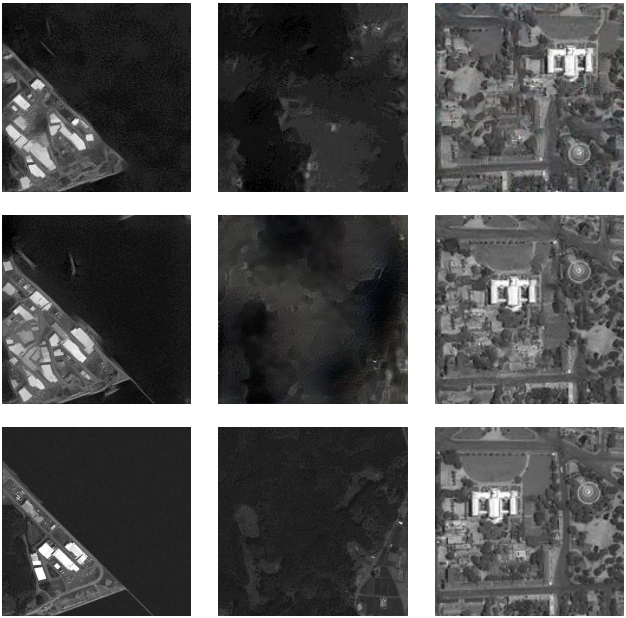

Fig. 14. Transformed optical images with different pooling operations. As for the images above, different columns correspond to different datasets. (The first, the second, and the third columns are from the first, the second, and the third datasets, respectively.) Different rows represent different pooling operations. (The first and the second rows represents the images with average and max pooling operations, respectively.) The third row gives the real optical images for comparison.) As can be seen, the images with the max pooling operation are the most homogeneous with the optical images.

content features.

In conclusion, as the deepest convolutional layer with the powerful capability of representing the high-level features, Conv5-4 is selected to extract the semantic content.

## APPENDIX II

The max pooling operation, utilized in the VGG network shown in Figs. 1 and 2, is compared with the average pooling operation. In [28], the average pooling is preferred for natural images but no experimental comparison is presented with the max pooling. Fig. 14 shows the transformed images with different pooling operations on the experimental datasets.

In the first dataset (first column of Fig. 14), the narrow breakwater in the lower-right is lost in the image with the average pooling but preserved in the image with the max pooling operation. It means that the max pooling preserves the semantic content better. It holds true for the second dataset (second column of Fig. 14), in which the semantic content of the transformed image is damaged seriously: the farmland in the lower-right vanishes and several buildings appears in the wrong place (i.e., forest regions). In the third dataset (third column of Fig. 14), with the average pooling operation, the white building in the middle and the circle building in the upper-right are misplaced. By applying the max pooling operation, the two buildings are placed correctly, compared with the real optical image.

We also evaluate the detection results quantitatively in Table VI. Because of better preservation of semantic content, the max pooling derives the change detection results more accurately than the average pooling. Therefore, max pooling operation is employed in the VGG network to extract semantic content and style features.

TABLE VI
COMPARISONS OF DETECTION RESULTS BY DIFFERENT POOLING OPERATIONS (%)

| Pooling operation | First dataset | | Second dataset | | Third dataset | |
|---|---|---|---|---|---|---|
| | $R_a$ | $Ka$ | $R_a$ | $Ka$ | $R_a$ | $Ka$ |
| Average | 98.56 | 75.13 | 98.61 | 64.07 | 95.51 | 62.23 |
| Max | 98.63 | 76.00 | 98.93 | 73.77 | 96.23 | 64.04 |